\documentclass[runningheads]{llncs}

\usepackage{header}

\usepackage[utf8]{inputenc} %
\usepackage[T1]{fontenc}    %
\usepackage{cite}
\usepackage{amsmath,amssymb,amsfonts}
\usepackage{algorithmic}
\usepackage{graphicx}
\usepackage{textcomp}
\usepackage{xcolor}
\def\BibTeX{{\rm B\kern-.05em{\sc i\kern-.025em b}\kern-.08em
    T\kern-.1667em\lower.7ex\hbox{E}\kern-.125emX}}

\usepackage{multirow}
\usepackage{tabularx,ragged2e}
\usepackage{tabulary}
\usepackage{adjustbox}
\usepackage{arydshln}
\usepackage{hyperref}
\usepackage{cleveref}
\usepackage[flushleft]{threeparttable}
\usepackage{url}
\usepackage{hhline}
\usepackage{multicol}
\usepackage{booktabs}
\usepackage{footmisc}
\usepackage{graphicx}
\usepackage{pifont}
\usepackage[normalem]{ulem}
\usepackage{soul}
\usepackage{color}
\usepackage{wrapfig}
\usepackage{etoolbox}

\definecolor{lightgray}{rgb}{0.95,0.85,0.85}
\sethlcolor{lightgray}

\usepackage{arydshln}

\newcommand{\spinnerchiefif}{Spinner\-Chief-IF }
\newcommand{\spinnerchiefdf}{Spinner\-Chief-DF }

\definecolor{Gray}{gray}{0.925} % Table
\definecolor{Preprint}{rgb}{.63,.79,.95}
\definecolor{darkblue}{rgb}{0, 0, 0.5}
\hypersetup{colorlinks=true, citecolor=darkblue, linkcolor=darkblue, urlcolor=darkblue}

\begin{document}

\title{Identifying Machine-Paraphrased Plagiarism}

%=====================================================================================
% Condensed Authors
%=====================================================================================

\author{Jan Philip Wahle\inst{1}\orcidID{0000-0002-2116-9767} \and Terry Ruas\inst{1}\orcidID{0000-0002-9440-780X} \and Tom\'{a}\v{s} Folt\'{y}nek\inst{2}\orcidID{0000-0001-8412-5553} \and Norman Meuschke\inst{1}\orcidID{0000-0003-4648-8198} \and Bela Gipp\inst{1}\orcidID{0000-0001-6522-3019}}
%

% \authorrunning{J. P. Wahle et al.}
\authorrunning{ }

\institute{University of Wuppertal, Rainer-Gruenter-Straße, 42119 Wuppertal, Germany \email{last@uni-wuppertal.de} \\ \and Mendel University in Brno, 61300 Brno, Czechia \\ \email{tomas.foltynek@mendelu.cz}}

\maketitle

\thispagestyle{firststyle}

\begin{abstract}

Employing paraphrasing tools to conceal plagiarized text is a severe threat to academic integrity.
To enable the detection of machine-paraphrased text, we evaluate the effectiveness of five pre-trained word embedding models combined with machine-learning classifiers and eight state-of-the-art neural language models.
We analyzed preprints of research papers, graduation theses, and Wikipedia articles, which we paraphrased using different configurations of the tools SpinBot and SpinnerChief.
The best-performing technique, Longformer, achieved an average F1 score of 81.0\% (F1=99.7\% for SpinBot and F1=71.6\% for SpinnerChief cases), while human evaluators achieved F1=78.4\% for SpinBot and F1=65.6\% for SpinnerChief cases.
We show that the automated classification alleviates shortcomings of widely-used text-matching systems, such as Turnitin and PlagScan.
To facilitate future research, all data\footnote{\url{https://doi.org/10.5281/zenodo.3608000}}, code\footnote{\url{https://github.com/jpwahle/iconf22-paraphrase}}, and two web applications\footnote{\url{http://purl.org/spindetector}}\footnote{\url{https://huggingface.co/jpelhaw/longformer-base-plagiarism-detection}} showcasing our contributions are openly available.


\end{abstract}

%=====================================================================================
% Introduction
%===================================================================================== 
\section{Introduction}\label{sec.intro}
Plagiarism is a pressing problem for educational and research institutions, publishers, and funding agencies \cite{Foltynek2019}. 
To counteract plagiarism, many institutions employ \textit{text-matching software}. These tools reliably identify duplicated text yet are significantly less effective for paraphrases, translations, and other concealed forms of plagiarism \cite{Foltynek2019, testop2020}. 

Studies show that an alarming proportion of students employ \textit{online paraphrasing tools} to disguise text taken from other sources~\cite{Rogerson2017,prentice2018}.
These tools employ artificial intelligence approaches to change text, e.g., by replacing words with their synonyms \cite{zhang2014}.
Paraphrasing tools serve to alter the content so that search engines do not recognize the fraudulent websites as duplicates.

In academia, paraphrasing tools help to mask plagiarism, facilitate collusion, and help ghostwriters with producing work that appears original.
These tools severely threaten the effectiveness of text-matching software, which is a crucial support tool for ensuring academic integrity.
The academic integrity community calls for technical solutions to identify the machine-paraphrased text as one measure to counteract paraphrasing tools \cite{Rogerson2017}.

The International Journal for Educational Integrity recently devoted a special issue\footnote{\url{https://edintegrity.biomedcentral.com/mbp}} to this topic. 

We address this challenge by devising an automated approach that reliably distinguishes human-written from machine-paraphrased text and providing the solution as a free and open-source web application.

In this paper, we extend Folt\'{y}nek et al.\cite{Foltynek2020} work by proposing two new collections created from research papers on arXiv\footnote{\url{https://arxiv.org}} and graduation theses of ``English language learners'' (ELL), and explore a second paraphrasing tool for generating obfuscated samples. We also include eight neural language models based on the Transformer architecture for identifying machine-paraphrases.

%=====================================================================================
% Related Work
%===================================================================================== 
\section{Related Work}\label{sec.relwork}
The research on plagiarism detection technology has yielded many approaches that employ lexical, syntactical, semantic, or cross-lingual text analysis \cite{Foltynek2019}.
These approaches reliably find copied and moderately altered text; some can also identify paraphrased and machine-translated text. Methods to complement text analysis focus on non-textual features \cite{Meuschke2021}, such as academic citations \cite{Meuschke2018a}, images \cite{Meuschke2018}, and mathematical content \cite{Meuschke2019}, to improve the detection of concealed plagiarism. 

Most research on paraphrase identification quantifies to which degree the meaning of two sentences is identical.
Approaches for this task employ lexical, syntactic, and semantic analysis (e.g., word embedding) as well as machine learning and deep learning techniques~\cite{Foltynek2019,Wahle21b}.

The research on distinguishing machine-paraphrased text passages from original content is still in an early stage. Zhang et al.~\cite{zhang2014} provided a tool that determines if two articles are derived from each other.
However, they did not investigate the task of distinguishing original and machine-fabricated text. Dey et al.~\cite{dey2016} applied a Support Vector Machine (SVM) classifier to identify semantically similar tweets and other short texts.  A very recent work studied word embedding models for paraphrase sentence pairs with word reordering and synonym substitution \cite{Alvi21}. In this work, we focus on detecting paraphrases without access to pairs as it represents a realistic scenario without pair information.

Techniques to accomplish the task of paraphrase detection, dense vector representations of words in documents have attracted much research in recent years.
Word embedding techniques, such as word2vec~\cite{MikolovSCC13}, have alleviated common problems in bag-of-words (BOW) approaches, e.g., scalability issues and the curse of dimensionality.
Representing entire documents in a single fixed-length dense vector (doc2vec) is another successful approach~\cite{Le:14}.
Word2vec and doc2vec can both capture latent semantic meaning from textual data using efficient neural network language models.
Prediction-based word embedding models, such as word2vec and doc2vec, have proven themselves superior to count-based models, such as BOW, for several problems in Natural Language Processing (NLP), such as quantifying word similarity~\cite{RuasGA19}, classifying documents~\cite{RuasFGd20}, and analyzing sentiment \cite{Perone:18}.
Gharavi et al. employed word embeddings to perform text alignment for sentences~\cite{Gharavi2020}.
Hunt et al. integrated features from word2vec into machine learning models (e.g., logistic regression, SVM) to identify duplicate questions in the Quora dataset~\cite{Hunt2019}. We, on the other hand, consider text documents generated with the help of automated tools at the paragraph level.

Recently, the NLP community adapted and extended the neural language model BERT~\cite{DevlinCLT19} for a variety of tasks~\cite{BeltagyLC19a,YangDYC19, OstendorffRBG20, ClarkLLM20, OstendorffARG21a, Spinde2021f, Wahle22a}, similar to the way that word2vec \cite{MikolovSCC13} has influenced many later models in NLP~\cite{BojanowskiGJM17,RuasGA19,RuasFGd20}.
Based on the Transformer architecture~\cite{VaswaniSPU17}, BERT employs two pre-training tasks, i.e., \textit{Masked Language Model} (MLM) and \textit{Next Sentence Prediction} (NSP), to capture general aspects of language.
MLM uses a deep bidirectional architecture to build a language model by masking random tokens in the input.
The NSP task identifies if two sentences are semantically connected. The ALBERT~\cite{LanCGG19}, DistilBERT~\cite{SanhDCW19}, and RoBERTa~\cite{LiuOGD19} models are all based on BERT and either improve their predecessor's performance through hyperparameter adjustments or make BERT less computationally expensive.
Different from ELMo~\cite{PetersNIG18} and GPT~\cite{Radford2018}, BERT considers left-to-right and right-to-left context simultaneously, allowing a more realistic representation of the language. Although ELMo does use two LSTM networks, their weights are not shared during training. On top of MLM and NSP, BERT requires fine-tuning to specific tasks to adjust its weights accordingly. 

Other recent models proposed architectural and training modifications for BERT.
ELECTRA changes BERT's MLM task to a generator-discriminator setup~\cite{ClarkLLM20}. Tokens are substituted with artificially generated ones from a small masked language model and discriminated in a noise contrastive learning process~\cite{GutmannH10}.
BART pre-trains a bidirectional auto-encoding and an auto-regressive Transformer in a joint structure~\cite{LewisLGG19}.
The two-stage denoising auto-encoder first corrupts the input with an arbitrary function (bidirectional) and uses a sequence-to-sequence approach to reconstruct the original input (auto-regressive)~\cite{LewisLGG19}.
In XLNet, a permutation language modeling predicts one word given its preceding context at random~\cite{YangDYC19}.
Longformer proposed the most innovative contribution by exploring a new scheme for calculating attention~\cite{BeltagyPC20}.
Longformer's attention mechanism combines windowed local with global self-attention while also scaling linearly with the sequence length compared to earlier models (e.g., RoBERTa).

Folt\'{y}nek et al.~\cite{Foltynek2020} tested the effectiveness of six word embedding models and five traditional machine learning classifiers for identifying machine-paraphrased.
We paraphrased Wikipedia articles using the {SpinBot}\footnote{\url{https://spinbot.com/}\label{fn_spinbot}} API, which is the technical backbone of several widely-used services, such as Paraphrasing Tool\footnote{\url{https://paraphrasing-tool.com/}} and Free Article Spinner\footnote{\url{https://free-article-spinner.com/}}~\cite{Rogerson2017}.
The limitations of \cite{Foltynek2020} are the exclusive use of one data source, the lack of recent neural language models, and the reliance on a single paraphrasing tool.
In this paper, we address all three shortcomings by considering arXiv and graduation theses as new data sources (Section~\ref{sec.datasets}), eight neural language models (Section~\ref{sec.neuralapproaches}), and {SpinnerSchief}\footnote{\url{http://www.spinnerchief.com/}\label{fn_spinnerchief}} as an additional paraphrasing tool (Section~\ref{sec.opt}).  

Lan et al.~\cite{LanX18} compared five neural models (e.g., LSTM and CNN) using eight NLP datasets, of which three focus on sentence paraphrase detection (i.e., Quora~\cite{IyerDC17}, Twitter-URL~\cite{LanQHX17}, and PIT-2015~\cite{xu2014data}).
Subramanian et al. presented a model that combines language modeling, machine translation, constituency parsing, and natural language inference in a multi-task learning framework for sentence representation~\cite{SubramanianTBP18}.
Their model produces state-of-the-art results for the MRPC~\cite{DolanB05} dataset.
Our experiments consider a multi-source paragraph-level dataset and more recent neural models to reflect a real-world detection scenario and investigate recent NLP techniques that have not been investigated for this use case before. 

Wahle et al.~\cite{Wahle21b} is the only work, to date, that applies neural language models to generate machine paraphrased text.
They use BERT and other popular neural language models to paraphrase an extensive collection of original content. 
We plan to investigate additional models and combine them with the work on generating paraphrased data \cite{Wahle21b}, which could be used for training.

%=====================================================================================
% Methodology
%===================================================================================== 
\section{Methodology}\label{sec.method}

Our primary research objective is to provide a free service that distinguishes human-written from machine-paraphrased text while being insensitive to the topic and type of documents and the paraphrasing tool used.
We analyze paragraphs instead of sentences or entire documents since it represents a more realistic detection task \cite{Rogerson2017,weberwulff2019}. Sentences provide little context and can lead to more false positives when sentence structures are similar. Fulltext documents are computationally expensive to process, and in many cases the extended context does not provide a significant advantage over paragraphs.
We extend Folt\'{y}nek et al.'s~\cite{Foltynek2020} study by analyzing two new datasets (arXiv and theses), including an extra machine-paraphrasing tool (SpinnerChief), and evaluating eight state-of-the-art neural language models based on Transformers~\cite{VaswaniSPU17}.
We first performed preliminary experiments with classic machine learning approaches to identify the best-performing baseline methods for paraphrasing tools and datasets we investigate.
Next, we compared the best-performing machine learning techniques to neural language models based on the Transformer architecture, representing the latest advancements in NLP.

%-------------------------------------------------------------------------------------
% Paraphrasing Tools
%-------------------------------------------------------------------------------------
\subsection{Paraphrasing Tools}\label{sec.opt}
We employed two commercial paraphrasing services, i.e., \textit{SpinBot}\footref{fn_spinbot} and \textit{SpinnerChief}\footref{fn_spinnerchief}, to obfuscate samples in our training and test sets.
We used SpinBot to generate the training and test sets and SpinnerChief only for the test sets. 

SpinnerChief allows specifying the ratio of words it tries to change.
We experimented with two configurations: the \textit{default frequency (\spinnerchiefdf)}, which attempts to change every fourth word, and an \textit{increased frequency (\spinnerchiefif)}, which attempts to change every second word.

%-------------------------------------------------------------------------------------
% Datasets for Training and Testing
%-------------------------------------------------------------------------------------
\subsection{Datasets for Training and Testing}\label{sec.datasets}
Most paraphrasing tools are paid services, which prevents experimenting with many of them.
The financial costs and effort required for obtaining and incorporating tool-specific training data would be immense.
Therefore, we employed transfer learning, i.e., used pre-trained word embedding models, trained the classifiers in our study on samples paraphrased using SpinBot, and tested whether the classification approach can also identify SpinnerChief's paraphrased text. 

\paragraph{Training Set:} We reused the paragraph training set of Folt\'{y}nek et al.~\cite{Foltynek2020} and paraphrased all 4,012 \textit{featured articles} from English Wikipedia using SpinBot\footref{fn_spinbot}.
We chose featured Wikipedia articles because they objectively cover a wide range of topics in great breadth and depth\footnote{\url{https://en.wikipedia.org/wiki/Wikipedia:Content_assessment}\label{fn_art-quality}}.
Approx. 0.1\% of all Wikipedia articles carry the label \textit{featured article}.

Thus, they are written in high-quality English by many authors and unlikely to be biased towards individual writing styles. 

The training set comprises of 200,767 paragraphs (98,282 original, 102,485 paraphrased) extracted from 8,024 Wikipedia articles.
We split each Wikipedia article into paragraphs and discarded those with fewer than three sentences, as Folt\'{y}nek et al.~\cite{Foltynek2020} showed such paragraphs often represent titles or irrelevant information.

%test pipeline
\paragraph{Test Sets:} Our study uses three test sets that we created from preprints of research papers on arXiv, graduation theses, and Wikipedia articles.
\Cref{tab:test_sets} summarizes the test sets.
For generating the \textbf{\textit{arXiv}} test set, we randomly selected 944 documents from the \textit{no problems} category of the arXMLiv project\footnote{\url{https://kwarc.info/projects/arXMLiv/}}.
The \textbf{\textit{Wikipedia}} test set is identical to~\cite{Foltynek2020}.
The paragraphs in the test set were generated analogously to the training set.
The \textbf{\textit{theses}} test set comprises paragraphs in 50 randomly selected graduation theses of ELL at the Mendel University in Brno, Czech Republic.
The theses are from a wide range of disciplines, e.g., economics, computer science, and cover all academic levels.
Unlike the arXiv and Wikipedia documents, the theses were only available as PDF files, thus required conversion to plain text.
We removed all content before the introduction section of each thesis, the bibliography, and all appendices to avoid noisy data (e.g., table of contents).

%Dataset details
% \usepackage{graphicx}
\begin{table}[htb]
\caption{Overview of the test sets.}
\label{tab:test_sets}
\centering
\scalebox{0.9}{%
\begin{threeparttable}
\begin{tabular}{lrrrrrr}
\toprule
\textbf{No. of paragraphs} & \multicolumn{2}{c}{\textbf{arXiv}}                                                                         & \multicolumn{2}{c}{\textbf{theses}}                                                                        & \multicolumn{2}{c}{\textbf{Wikipedia}}                                                                     \\
                           & \multicolumn{1}{c}{Original} & \multicolumn{1}{c}{\begin{tabular}[c]{@{}c@{}}Para-\\ phrased\end{tabular}} & \multicolumn{1}{c}{Original} & \multicolumn{1}{c}{\begin{tabular}[c]{@{}c@{}}Para-\\ phrased\end{tabular}} & \multicolumn{1}{c}{Original} & \multicolumn{1}{c}{\begin{tabular}[c]{@{}c@{}}Para-\\ phrased\end{tabular}} \\ \midrule
\textbf{SpinBot}           & 20,966                       & 20,867                                                                      & 5,226                        & 3,463                                                                       & 39,261                       & 40,729                                                                      \\

\textbf{SpinnerChief-DF}   & 20,966                       & 21,719                                                                      & 2,379                        & 2,941                                                                       & 39,261                       & 39,697                                                                      \\ 
\textbf{SpinnerChief-IF}   & 20,966                       & 21,671                                                                      & 2,379                        & 2,941                                                                       & 39,261                       & 39,618                                                                      \\ \bottomrule
\end{tabular}%
%TR: notes about training in spinbot removed
\end{threeparttable}
}
\end{table}

%-------------------------------------------------------------------------------------
% Word Embedding Models
%-------------------------------------------------------------------------------------
\subsection{Word Embedding Models}\label{sec.method_embeddings}
\Cref{tab:model-details} summarizes the word embedding models analyzed in our experiments: GloVe\footnote{\url{https://nlp.stanford.edu/projects/glove/}}~\cite{Penni:14}, word2vec\footnote{\url{https://code.google.com/archive/p/word2vec/}}(w2v)~\cite{MikolovSCC13}, fastText\footnote{\url{https://fasttext.cc/docs/en/english-vectors.html}}(FT-rw and FT-sw)~\cite{BojanowskiGJM17}, and doc\-2\-vec (d2v)~\cite{Le:14} that we trained from scratch. The d2v model uses a distributed bag-of-words training objective, a window size of 15 words, a minimum count of five words, trained word-vectors in skip-gram fashion, averaged word vectors, and 30 epochs. All word embedding models have 300 dimensions. Parameters we do not explicitly mention correspond to the default values in the \textit{gensim} API\footnote{\url{https://radimrehurek.com/gensim/models/doc2vec.html}}.

Our rationale for choosing the pre-trained word embedding models was to explore the most prominent techniques regarding their suitability for the plagiarism detection task. GloVe \cite{Penni:14} builds a co-occurrence matrix of the words in a corpus and explores the word probabilities ratio in a text to derive its semantic vectors as a count-based model. The training of w2v tries to predict a word given its context (cbow) or the context given a word (skip-gram) \cite{MikolovSCC13}. 

Even though numerous NLP tasks routinely apply GloVe and w2v \cite{conneau2017,RuasGA19,RuasFGd20}, they do not consider two important linguistic characteristics: word ordering and sub-wording. To explore these characteristics, we also included fastText \cite{BojanowskiGJM17} and the paragraph vector model \cite{Le:14}. 
FastText builds its word representation by extending the skip-gram model with the sum of the n-grams of its constituent sub-word vectors. As the paraphrasing algorithms used by plagiarists are unknown, we hypothesize rare words can be better recognized by fastText through sub-wording.
Two training options exist for the d2v model---Distributed Memory Model of Paragraph Vectors (pv-dm) and Distributed Bag of Words version of Paragraph Vector (pv-dbow). 
The former is akin to w2v cbow, while the latter is related to w2v skip-gram. Both options introduce a new paragraph-id vector that updates each context window on every timestamp. 
The paragraph-id vector seeks to capture the semantics of the embedded object. We chose a pv-dbow over a pv-dm model because of its superior results in semantic similarity tasks \cite{Lau:16}.

\begin{table}[htb]

\centering
\caption{Word embedding models in our experiments.}     \label{tab:model-details}
\scalebox{0.9}{
\begin{threeparttable}
\begin{tabular}{lll}\toprule
\textbf{Algorithm} & \textbf{Main Characteristics}                      & \textbf{Training Corpus}                     \\ \midrule
GloVe     & Word-word co-occurrence matrix            & Wikipedia Dump 2014 + Gigaword 5    \\
word2vec  & Continuous Bag-of-Words                   & Google News                        \\
pv-dbow   & Distributed Bag-of-Words    &Wikipedia Dump 2010 \\
fastText-rw  & Skip-gram without sub-words                                  & Wikipedia Dump 2017 + UMBC     \\
fastText-sw  & Skip-gram with sub-words                                 & Wikipedia Dump 2017 + UMBC         \\ \bottomrule
\end{tabular}
%TR 300 dimensions is in the text already
\end{threeparttable}
      }
\end{table} 

In our experiments, we represented each text as the average of its constituent word vectors by applying the word embedding models in \Cref{tab:model-details} \cite{ruas2019,RuasFGd20}. All models, except for d2v, yield a vector representation for each word. 
D2v produces one vector representation per document. 
Inferring the vector representations for unseen texts requires an additional training step with specific parameter tuning. We performed this extra training step with hyperparameters according to \cite{RuasFGd20} for the \textit{gensim} API: $\alpha = 10^{-4}$, $\alpha_{min}=10^{-6}$, and 300 epochs \cite{Lau:16}. The resulting pv-dbow embedding model requires at least 7 GB of RAM, compared to 1-3 GB required for other models. The higher memory consumption of pv-dbow can make it unsuitable for some use cases. 

%-------------------------------------------------------------------------------------
% Machine Learning Classifiers
%-------------------------------------------------------------------------------------
\subsection{Machine Learning Classifiers}\label{sec.classifiers}
After applying the pre-trained models to our training and test sets, we passed on the results to three machine learning classifiers: Logistic Regression (LR), Support Vector Machine (SVM), and Na\"{i}ve Bayes (NB).

We employed a grid-search approach implemented using the scikit-learn package\footnote{\url{https://scikit-learn.org}} in Python for finding the optimal parameter values for each classifier (\Cref{tab:gridsearch})

\begin{table}[ht!]
\small
\centering
\caption{Grid-search parameters for ML classifiers.} \label{tab:gridsearch}
  \begin{tabular}{llr}
  \toprule
    Classifier                                                                        & Parameter             & Range                                       \\\midrule
    %\\\cmidrule(lr){2-3}
    \multirow{4}{*}{\begin{minipage}{0.7in}Logistic\\ Regression\end{minipage}}       & solver                & newton-cg, lbfgs, sag, saga                 \\
                                                                                      & maximum iteration     & 500, 1000, 1500                             \\
                                                                                      & multi-class           & ovr, multinomial                            \\
                                                                                      & tolerance             & 0.01, 0.001, 0.0001, 0.00001                \\\cmidrule(lr){2-3}
    \multirow{4}{*}{\begin{minipage}{0.5in}Support\\ Vector\\Machine\end{minipage}}   & kernel                & linear, radial bases function, polynomial   \\
                                                                                      & gamma                 &  0.01, 0.001, 0.0001, 0.0001                \\
                                                                                      & polynomial degree     & 1, 2, 3, 4, 5, 6, 7, 8, 9                   \\
                                                                                      & C                     & 1, 10, 100                                  \\
    \bottomrule
  \end{tabular}
\end{table}

%\footnotetext{https://scikit-learn.org}

%-------------------------------------------------------------------------------------
% Neural Language Model
%-------------------------------------------------------------------------------------
\subsection{Neural Language Models}\label{sec.neuralapproaches}
We investigate the following neural language models based on the Transformer architecture: BERT~\cite{Devlin:18}, RoBERTa\cite{Liu2019}, ALBERT~\cite{Lang2019}, DistilBERT~\cite{SanhDCW19}, ELECTRA~\cite{ClarkLLM20}, BART~\cite{LewisLGG19}, XLNet~\cite{YangDYC19}, and Longformer~\cite{BeltagyPC20}.
Our rationale for testing Transformer-based models is their ability to generally outperform traditional word embedding and machine learning models in similar NLP tasks (e.g., document similarity). We chose the aforementioned models specifically because of two reasons. First, we explore models closely related or based on BERT that improve BERT through additional training time and data (RoBERTa) or compress BERT's architecture with minimal performance loss (DistilBERT, AlBERT). Second, we use contrasting models to BERT that, although relying on the Transformer architecture, significantly change the training objective (XLNet), the underlying attention mechanism (Longformer), or employ a discriminative learning approach (ELECTRA, BART).

To classify whether a paragraph is paraphrased, we attach a randomly initialized linear layer on top of the model's embedding of the aggregate token (e.g., \texttt{[CLS]} for BERT) to transform the embedding into binary space. The final layer predicts whether a paragraph has been paraphrased using cross-entropy loss. For all models, we use the base version, the official pre-trained weights, and the following configurations: a sequence length of $512$ tokens, an accumulated batch size of $32$, the Adam optimizer with $\alpha = 2e-5$, $\beta_1 = 0.9$, $\beta_2 = 0.999$, $\epsilon = 1e-8$, and PyTorch's native automated mixed-precision format. Using a common sequence length of 512 tokens allows for a fair comparison of the models without losing important context information\footnote{99.35\% of the datasets' text can be represented with less than 512 tokens.}. 
\Cref{sec.eval_nlm} provides more details about the models' characteristics.

%=====================================================================================
% Evaluation
%===================================================================================== 
\section{Evaluation}\label{sec.eval}
To quantify the effectiveness of classification approaches in identifying machine-paraphrased text, we performed three experiments. \Cref{sec.eval_auto} presents the results of applying the pre-trained word embedding models in combination with machine learning classifiers and neural language models to the three test sets. \Cref{sec.eval_human,sec.eval_turnitin} indicate how well human experts and a text-matching software identify machine-paraphrased text to put the results of automated classification approaches into context.

%-------------------------------------------------------------------------------------
% Automated classification
%-------------------------------------------------------------------------------------
\subsection{Automated Classification}\label{sec.eval_auto}
This section presents the micro-averaged F1 scores (\mbox{F1-Micro}) for identifying paragraphs we paraphrased using either SpinBot or SpinnerChief and classified using combinations of pre-trained word embedding models and machine learning classifiers or Transformer-based language models.

%-------------------------------------------------------------------------------------
% ML + SpinBot
%-------------------------------------------------------------------------------------
\paragraph{Results of ML Techniques for SpinBot:}\label{sec.eval_ml_sb} 

\Cref{tab:class-results-sb} shows the results for classifying SpinBot test sets derived from arXiv, theses, and Wikipedia using combinations of pre-trained word embedding models and machine learning classifiers. GloVe, in combination with SVM, achieved the best classification performance for all test sets. The combination of w2v and SVM performed nearly as well as GloVe+SVM for all test sets. For the theses and Wikipedia test sets, the performance difference between GloVe+SVM and w2v+SVM is less than 2\%, and for the arXiv test set 6.66\%. All pre-trained word embedding models achieved their best results for the Wikipedia test set. 

\begin{table*}[htb]
\label{tab:class-results-sb}
\small
\centering
\begin{threeparttable}
\centering
\caption{Classification results (F1-Micro) of ML techniques for SpinBot.} 
\begin{tabular}{lp{2cm}p{2cm}p{2cm}p{2cm}c}\toprule
                 & GloVe    & w2v      &  d2v       & FT-rw     & FT-sw    \\\midrule
\textbf{arXiv}           &\textbf{86.46}	&79.80	&72.40	&78.40	&74.14  \\ \midrule
LR              &76.53	&74.82	&69.42	&75.08	&65.92  \\
SVM	            &86.46	&79.80	&72.40	&76.31	&74.15	\\
NB	            &79.17	&74.23	&57.99	&78.40	&64.96	\\\midrule

\textbf{theses}          &\textbf{83.51}	&81.94	&61.92	&72.75	&64.78  \\ \midrule
LR              &68.55	&72.89	&59.97	&69.17	&64.03  \\
SVM	            &83.51	&81.94	&61.92	&72.75	&64.78	\\
NB	            &75.22	&74.18	&42.30	&72.11	&61.99	\\\midrule

\textbf{Wikipedia}       &\textbf{89.55}	&87.27	&83.04	&86.15	&82.57  \\ \midrule
LR              &80.89	&84.50	&81.08	&85.13	&78.97  \\
SVM	            &89.55	&87.27	&83.04	&86.15	&82.57	\\
NB	            &69.68	&69.84	&58.88	&70.05	&64.47	\\
\bottomrule
\end{tabular}
\begin{tablenotes}[online]
\item[\ding{227}] \textbf{Boldface} indicates the best score for each test set, i.e., arXiv, theses, and Wikipedia. The score of the best-performing combination of embedding model and classifier is repeated in the row of the test set. 
\end{tablenotes}
\end{threeparttable}
\end{table*}

All classification approaches, except for w2v+SVM, performed worst for the theses test set. However, the drop in performance for theses test cases is smaller than we expected. The F1-Micro score of the best approach for the theses test set (GloVe+SVM) is 6.04\% lower than for the Wikipedia test set and 3.09\% lower than for the arXiv test set. This finding suggests that the quality of writing in student theses mildly affects the detection of machine-paraphrased text.  

Although d2v seeks to mitigate shortcomings of its predecessor w2v, such as ignoring word order and producing a variable-length encoding, w2v surpassed d2v for all test sets. A possible reason is the short length of the paragraphs we consider. Lau et al. found that \mbox{d2v's} performance decreases for short documents \cite{Lau:16}. The results for paragraphs in \Cref{tab:class-results-sb} and for documents in our preliminary study~\cite{Foltynek2020}, where d2v was the best-performing approach, support this conclusion. 

For fastText (FT-rw and FT-sw in \Cref{tab:class-results-sb}), we observe the same behavior as for w2v and d2v. The sub-word embeddings of FT-sw should provide a benefit over FT-rw, which encodes whole words, by capturing sub-word structures~\cite{BojanowskiGJM17}. Therefore, we expected a better performance of FT-sw compared to FT-rw. However, FT-rw and simpler models, i.e., GloVe and w2v, performed better than FT-sw for all test sets. 

%-------------------------------------------------------------------------------------
% ML + SpinnerChief
%-------------------------------------------------------------------------------------
\paragraph{Results of ML Techniques for SpinnerChief:}\label{sec.eval_ml_sc}

\Cref{tab:class-results-sc-all} shows the results of applying the pre-trained word embedding models and machine learning classifiers to the arXix, theses, and Wikipedia test sets containing cases paraphrased by the SpinnerChief tool. We either used the tool's default setting, i.e., attempting to replace every fourth word (\spinnerchiefdf), or increased the frequency of attempted word replacements to every other word (\spinnerchiefif).

\begin{table*}[hbt]
\small
\centering
\caption{Classification results (F1-Micro) of ML techniques for SpinnerChief.} \label{tab:class-results-sc-all}
\begin{threeparttable}[ht!]
\begin{tabular}{lcccccccccc}\toprule
\multirow{2}{*}{} &\multicolumn{5}{c}{\textbf{SpinnerChief-DF}} & \multicolumn{5}{c}{\textbf{SpinnerChief-IF}}  \\
\cmidrule(lr){2-6}
\cmidrule(lr){7-11}
                & GloVe    & w2v      &  d2v       & FT-rw     & FT-sw    & GloVe    & w2v      &  d2v       & FT-rw     & FT-sw\\
\cmidrule(lr){1-6}
\cmidrule(lr){7-11}

\textbf{arXiv}           &58.48	&\textbf{59.78}	&56.46	&57.42	&59.72   &64.34	&\textbf{65.89}	&59.27	&63.70	&63.66  \\ 

\cmidrule(lr){1-6}
\cmidrule(lr){7-11}
LR              &52.14	&55.43	&56.46	&57.42	&58.64  &54.92	&59.61	&59.07	&61.74	&61.57  \\
SVM	            &58.42	&57.65	&56.43	&56.43	&59.72	 &64.12	&62.77	&59.27	&62.97	&63.66  \\
NB	            &58.48	&59.78	&51.58	&51.58	&55.21	&64.34	&65.89	&52.21	&63.70	&59.33  \\ 
\cmidrule(lr){1-6}
\cmidrule(lr){7-11}

\textbf{theses}          &52.63	&53.60	&\textbf{59.09}	&53.08	&57.25   &58.57	&58.24	&\textbf{63.15}	&59.13	&61.27  \\ 
\cmidrule(lr){1-6}
\cmidrule(lr){7-11}
LR              &48.42	&53.60	&59.09	&52.51	&55.63   &52.08	&57.94	&62.88	&59.13	&60.65  \\
SVM	            &52.63	&51.54	&59.00	&53.08	&57.25	 &58.57	&57.78	&63.15	&58.12	&61.27  \\
NB	            &50.90	&53.32	&54.94	&52.78	&46.99	&55.62	&58.24	&55.09	&57.19	&50.13  \\ 
\cmidrule(lr){1-6}
\cmidrule(lr){7-11}

\textbf{Wikipedia}       &57.86	&\textbf{60.30}	&55.99	&59.19	&59.62  &64.16	&\textbf{66.83}	&60.94	&65.35	&66.41  \\ 
\cmidrule(lr){1-6}
\cmidrule(lr){7-11}
LR              &52.97	&55.90	&55.64	&56.40	&59.62  &55.68	&61.32	&60.16	&62.51	&66.41  \\
SVM	            &57.09	&57.48	&55.99	&57.15	&58.72	&64.16	&64.56	&60.94	&63.61	&64.81  \\
NB	            &57.86	&60.30	&51.64	&59.19	&57.29	&63.46	&66.83	&52.64	&65.35	&62.06  \\ 
\bottomrule
\end{tabular}
\begin{tablenotes}[online]
\item[\ding{227}] \textit{-DF} default frequency, \textit{-IF} increased frequency (attempt changing every fourth or every second word). 
\item[\ding{227}] \textbf{Boldface} indicates the best score for each test set. The score of the best-performing combination of embedding model and classifier is repeated in the row of the test set.
\end{tablenotes}
\end{threeparttable}
\end{table*}

We observe a drop in the SpinnerChief's classification performance compared to the results for SpinBot. The average decrease in the F1-Micro scores was approx. 17\% when using \spinnerchiefdf and approx. 13\% for -IF. 
The comparison between the results of SpinBot and \spinnerchiefif is more informative than comparing SpinBot to \spinnerchiefdf since the IF setting yields a similar ratio of replaced words to SpinBot.

As in SpinBot, all approaches performed best for the Wikipedia and worst for the theses. However, the performance differences were smaller for SpinnerChief than for SpinBot. For all SpinnerChief (DF and IF), the lowest F1-Micro scores were at most 6.5\% below the highest scores, and the runner-ups were generally within an interval of 2\% of the best scores. 

The characteristics of ELL texts, e.g., sub-optimal word choice and grammatical errors, decreased the classification performance less than we had expected. The highest scores for the SpinBot theses are approx. 6\% lower than the highest scores for any other dataset for SpinBot. For SpinnerChief, this difference is approx. 2\%.

Notably, SpinnerChief's settings for a stronger text obfuscation (\spinnerchiefif) increased the rate with which the classification approaches identified the paraphrases.
On average, \spinnerchiefdf replaced 12.58\% and \spinnerchiefif 19.37\% of the words in the text (\Cref{sec.opt}).
The 6.79\% increase in the number of replaced words for \spinnerchiefif compared to \spinnerchiefdf increased the average F1-Micro score of the classification approaches by 5.56\%.
This correlation suggests that the classification approaches can recognize most of the characteristic word replacements that paraphrasing tools perform.

Text-matching software, such as Turnitin and PlagScan, are currently the de-facto standard technical support tools for identifying plagiarism. However, since these tools search for identical text matches, their detection effectiveness decreases when the number of replaced words increases (\Cref{tab:results-plag-turn}). Including additional checks, such as the proposed classification approaches, as part of the text-matching software detection process could alleviate the weaknesses of current systems.

We attribute the drop in the classification performance and the overall leveling of the F1-Micro scores for SpinnerChief test sets compared to SpinBot test sets to our transfer learning approach. As explained in \Cref{sec.method}, we seek to provide a system that generalizes well for different document collections and paraphrasing tools. Therefore, we used the machine-paraphrased text samples of SpinBot and applied the pre-trained word embedding models from \Cref{tab:model-details} to extract the vector representations. We then used these vectors as the features for the machine learning classifiers for both Spinbot and SpinnerChief test sets.

We selected the combinations of word embedding models and machine learning classifiers that performed best for SpinBot (\Cref{tab:class-results-sb}) and SpinnerChief (\Cref{tab:class-results-sc-all}) as the baseline to which we compare the Transformer-based language models in the following section.

%-------------------------------------------------------------------------------------
% ML + Transformer
%-------------------------------------------------------------------------------------
\paragraph{Results for Transformer-based Language Models:}\label{sec.eval_nlm}

\begin{table*}[t]
\small
\centering

    \begin{threeparttable}
\caption{Classification results (F1-Micro) of best ML techniques and neural language models for SpinBot and SpinnerChief.} \label{tab:class-results-nlm}
    \begin{tabular}{l c ccc c ccc c ccc} 
    \toprule
    \multirow{2}{*}{Techniques} & {} & \multicolumn{3}{c}{\textbf{SpinBot}} & {} & \multicolumn{3}{c}{\textbf{SpinnerChief-DF}} & {} & \multicolumn{3}{c}{\textbf{SpinnerChief-IF}}  \\
    
    \cmidrule(lr){3-5}
    \cmidrule(lr){7-9}
    \cmidrule(lr){11-13}
    & {} & arXiv & Theses & Wiki  & {} & arXiv & Theses & Wiki & {} & arXiv & Theses & Wiki           \\
    \toprule
    % \cmidrule(lr){3-5}
    % \cmidrule(lr){7-9}
    % \cmidrule(lr){11-13}
    Baseline                   & {} & 86.46\tnote{a} & 83.51\tnote{a}  & 89.55\tnote{a}     & {} & 59.78\tnote{b} & 59.09\tnote{c}  & 60.30\tnote{b}              & {} & 65.89\tnote{b} & 63.15\tnote{d}  & 66.83\tnote{b}               \\
    [0.5ex] \hdashline\noalign{\vskip 0.5ex}
    BERT               & {} & 99.44 & 94.72  & 99.85      & {} & 50.74 & 50.42  & 43.00                & {} & 64.59 & 63.59  & 57.45               \\
    ALBERT                & {} & 98.91 & 96.77  & 99.54     & {} & 66.88 & 47.92  & 50.43             & {} & 75.57 & 56.75  & 59.61               \\
    DistilBERT             & {} & 99.32 & 96.61  & 99.42      & {} & 38.37 & 45.07  & 37.05              & {} & 47.25 & 51.44  & 46.81               \\
    RoBERTa                     & {} & 99.05 & 97.34  & 99.85     & {} & 57.10  & 47.40 & 48.03             & {} & 66.00    & 58.24  & 58.94               \\
    [0.5ex] \hdashline\noalign{\vskip 0.5ex}
    ELECTRA                   & {} & 99.20  & 96.85  & 99.41      & {} & 43.83 & 44.95  & 56.30               & {} & 60.77 & 63.11  & \textbf{75.92}               \\
    BART                   & {} & 99.58 & 99.66  & 99.86      & {} & 69.38 & 53.39  & 48.62              & {} & 76.07 & 63.57  & 58.34               \\
    XLNet                       & {} & \textbf{99.65} & 98.33  & 99.48      & {} & 69.90  & 53.06  & 50.51             & {} & \textbf{80.56} & 71.75  & 61.83               \\
    Longformer                  & {} & 99.38 & \textbf{99.81}  & \textbf{99.87}      & {} & \textbf{76.44} & \textbf{70.15}  & \textbf{63.03 }             & {} & 78.34 & \textbf{74.82}  & 67.11               \\
    \bottomrule
    \end{tabular}
    \begin{tablenotes}[online]
        %\item[\ding{227}] \textit{-DF} default frequency, \textit{-IF} increased frequency (attempt changing every fourth or every second word).
        \item[\ding{227}]\textsuperscript{a}GloVe+SVM \hspace{0.4cm} \textsuperscript{b}w2v+NB  \hspace{0.4cm} \textsuperscript{c}d2v+LR \hspace{0.4cm} \textsuperscript{d}d2v+SVM
        \item[\ding{227}] The first horizontal block shows the best results of machine learning techniques, the second of models that optimize BERT, and the third of models that use new architectural or training approaches.
        \item [\ding{227}] \textbf{Boldface} indicates the best score for each test set.
    \end{tablenotes}
    \end{threeparttable}
\end{table*}

\Cref{tab:class-results-nlm} shows the classification results of neural language models applied to all SpinBot and SpinnerChief test sets. The machine learning technique that performed best for each test set (\Cref{tab:class-results-sb,tab:class-results-sc-all}) is shown as \textit{Baseline}.  

For the SpinBot, all Transformer-based models outperformed their machine learning counterparts on average by 16.10\% for theses, 13.27\% for arXiv, and 10.11\% for Wikipedia. Several models consistently achieved F1-Micro scores above 99\% for all SpinBot cases. These findings show that the models could capture the intrinsic characteristics of SpinBot's paraphrasing method very well. We stopped the training for each model after one epoch to avoid overfitting.

All techniques performed worse for SpinnerChief than for SpinBot, which we expected given the transfer learning approach. The drop in the classification performance was consistently lower for \spinnerchiefif, which exhibits a similar ratio of replaced words as SpinBot, than for \spinnerchiefdf, which contain fewer replaced words than SpinBot. The most significant improvements in the scores for \spinnerchiefif over \spinnerchiefdf are 16.94\% for arXiv (ELECTRA), 18.69\% for the theses (XLNet), and 19.62\% for Wikipedia (BART).

These results show that the ratio of replaced words is a significant indicator of a models' performance. However, since the paraphrasing methods of SpinBot and SpinnerChief (DF and IF) are unknown and could be different for each setting, one can interpret this finding in two ways. First, the models may capture the frequency of replaced words intrinsically and increase their attention to more words, which would mean the models can better detect more strongly altered paragraphs, such as those produced by \spinnerchiefif. Second, \spinnerchiefif cases might be better detectable because the paraphrasing method associated with the \spinnerchiefif setting might be more akin to the one of SpinBot than the method associated with the \spinnerchiefdf setting. 
 
For all \spinnerchiefdf cases, Longformer consistently achieved the best results, surpassing the F1-Micro scores of the machine learning baselines by 10.15\% on average and 16.66\% for arXiv. For \spinnerchiefif, XLNet, Longformer, and ELECTRA achieved the best results with an improvement in the F1-Micro scores of 14.67\%, 11.67\%, and 9.09\% over the baseline scores for the arXiv, theses, and Wikipedia, respectively. As ELECTRA was pre-trained using a Wikipedia dump and the Books Corpus~\cite{ZhuICCV2015}, we assume it also captured semantic aspects of Wikipedia articles. 

The larger diversity in the training data of Longformer and XLNet (i.e., Gigaword 5~\cite{NapolesMB2012}, CC Stories~\cite{TrinhL2019}, and Realnews~\cite{Zellers2019}) seems to enable the models to capture unseen semantic structures in the arXiv and theses better than other models.

BERT and its derived models performed comparably to the baselines for most SpinnerChief cases. DistilBERT, which uses knowledge distillation to reduce the number of parameters by 40\% compared to BERT, performed significantly worse than its base model. For the SpinnerChief, the F1-micro scores for DistilBERT were on average 10.63\% lower than for BERT, often falling into a score range achievable by random guessing. Although we expected a slight decline in the accuracy of DistilBERT compared to BERT due to the parameter reduction, the results fell well below our predictions. In comparison, for the General Language Understanding (GLUE) dataset~\cite{WangSMH19}, DistilBERT performed only 2.5\% worse than BERT. ALBERT's parameter reduction techniques, e.g., factorized embedding parametrization and parameter sharing, seem to be more robust. ALBERT outperformed BERT on average by 4.56\% on the SpinnerChief test sets. With an average improvement of 0.99\%, RoBERTa performed slightly better than BERT. However, as RoBERTa uses more parameters than most other BERT-related models and has exceptionally high computational requirements for pre-training, we rate this performance benefit as negligible.

In summary, improvements of BERT's attention mechanism or training objective outperformed other BERT-based models for the machine-paraphrase detection task. We hypothesize the windowed local and global self-attention scheme used in Longformer allowed the model to generalize better between different paraphrasing tools. In eight out of nine scenarios, Longformer was either the best or second-best model overall. Also, for almost all cases, the neural language models surpassed the machine learning approaches' results, thus providing a better solution to the problem of identifying machine-paraphrases. We see the \spinnerchiefdf test results set as a lower bound regarding the detection effectiveness for unseen spinning methods, even if the frequency of word replacements is significantly different from the frequency in our training set. 

%-------------------------------------------------------------------------------------
% Human Baseline
%-------------------------------------------------------------------------------------
\subsection{Human Baseline}\label{sec.eval_human}
To complement the earlier study of Folt\`{y}nek \cite{Foltynek2020}, we conducted a user survey with excerpts from ten randomly selected Wikipedia articles. We paraphrased three of the ten excerpts using SpinerChief-DF, three others using \spinnerchiefif, and kept four excerpts unaltered. Using QuizMaker\footnote{\url{https://www.quiz-maker.com/}}, we prepared a web-based quiz that showed the ten excerpts one at a time and asked the participants to vote whether the text had been machine-paraphrased. We shared the quiz via e-mail and a Facebook group with researchers from the academic integrity community and 32 participants joined our study. 

The participants' accuracy ranged between 20\% and 100\%, with an average of 65.59\%. Thus, the F1-Micro score of the 'average' human examiner matched the average of the best scores of automated classification approaches for the SpinnerChie-IF test sets (65.29\%). Some participants pointed out that oddness in the text of some excerpts, e.g., lowercase letters in acronyms, helped them identify the excerpts as paraphrased. For SpinBot, 73 participants answered the survey with an accuracy ranging from 40\% to 100\% (avg. 78.40\%) according to\cite{Foltynek2020}.  

Our experiments show that experienced educators who read carefully and expect to encounter machine-paraphrased text could achieve an accuracy between 80\% and 100\%. However, even in this setting, the average accuracy was below 80\% for SpinBot and below 70\% for SpinnerChief. We expect that the efficiency will be lower in a realistic scenario, in which readers do not pay special attention to spotting machine paraphrases. 

%-------------------------------------------------------------------------------------
% PDS Baseline
%-------------------------------------------------------------------------------------
\subsection{Text-matching Software Baseline}\label{sec.eval_turnitin}
To quantify the benefit of our automated classification over text-matching software, we tested how accurately current text-matching tools identify paraphrased text. We tested two systems---\mbox{Turnitin}, which has the largest market share, and \mbox{PlagScan}, which was one of the best-performing systems in a comprehensive test conducted by the European Network for Academic Integrity (ENAI)~\cite{testop2020}. Our main objective was to test the tools' effectiveness in identifying patchwriting, i.e., inappropriately paraphrasing copied text by performing minor changes and substitutions. Patchwriting is a frequent form of plagiarism, particularly among students. 

For this test, we created four sets of 40 documents each (160 documents total). We composed each document by randomly choosing 20 paragraphs from Wikipedia articles (2x40 documents), arXiv preprints (40 documents), and theses (40 documents). For each set of 40 documents, we followed the same scheme regarding the length and obfuscation of the chosen paragraphs. First, we created ten documents by varying the paragraphs' length taken from the source from one to ten sentences. In addition to using the ten documents unaltered, we also paraphrased all ten documents using SpinBot, \spinnerchiefdf, and \spinnerchiefif.

\begin{table}[htb]
\caption{Percentage of text overlap reported by the text-matching systems Turnitin and PlagScan} \label{tab:results-plag-turn}
    \small
    \centering
    \scalebox{0.9}{
        \begin{tabular}{llccc}\toprule
        Detection     &Corpus      & arXiv    & Theses      &  Wikipedia       \\
        \midrule
        
        \multirow{4}{*}{Turnitin}
        
        &Original                &84.0	&5.4	&98.7 \\
        &SpinBot                 &7.0	&1.1	&30.2 \\
        &SpinnerChief-DF         &58.5	&4.0	&74.5 \\
        &SpinnerChief-IF          &38.8	&1.2	&50.1 \\ \midrule
        
        \multirow{4}{*}{PlagScan}
        &Original                 &44.6	&22.3	&65.0 \\
        &SpinBot                  &0.0	&0.1	&0.5  \\
        &SpinnerChief-DF          &9.2	&12.0	&19.1 \\
        &SpinnerChief-IF          &1.8	&0.5	&3.1\\

        \bottomrule
        \end{tabular}
    }
\end{table}

To ensure this test is objective and comparable across the data sets, we exclusively report the overall percentages of matching text reported by a system (\Cref{tab:results-plag-turn}). In most cases, the systems identified the correct source. However, the systems often reported false positives caused by random matches, which means the systems' actual retrieval effectiveness is slightly lower than reported.

The results show that PlagScan struggled to identify patchwriting. Even though the system indexes Wikipedia and could identify entirely plagiarized documents in the ENAI test~\cite{testop2020}, the average percentage of matching text reported for our patch-written documents was 63\%. Paraphrasing documents using SpinBot and \spinnerchiefif consistently prevented PlagScan from identifying the plagiarism. The average reported percentage of matching text was only 1\% for SpinBot and 3\% for \spinnerchiefif test cases. For \spinnerchiefdf test cases, PlagScan could identify 19\% of plagiarism present in documents, likely due to the lower ratio of altered words. Nevertheless, we can conclude obfuscating patch-written documents using a machine-paraphrasing tool likely prevents PlagScan from identifying plagiarism.

As shown in \Cref{tab:results-plag-turn}, Turnitin performed better for patch-written documents than PlagScan. 
For Wikipedia test cases, Turnitin reported 100\% matching text for almost all cases. 
The average percentage of matching text Turnitin reported for machine-paraphrased documents was much higher than for PlagScan---31\% for SpinBot, 74\% for \spinnerchiefdf, and 50\% for \spinnerchiefif.
However, machine-paraphrasing still prevents Turnitin from identifying a significant portion of the plagiarized content. 
Notably, Turnitin appears to index fewer theses than PlagScan, thus failing to report suspiciously high percentages of matching text for any theses test set, including ones containing unaltered paragraphs copied from theses.

For both systems, we observed the longer a plagiarized passage is, the more likely text-matching tools identified it. This result corresponds to the results of our classification approaches, which also yielded higher accuracy for longer passages.

From our experiments with text-matching software, we conclude that if plagiarists copy a few paragraphs and employ a paraphrasing tool to obfuscate their misconduct, the similarity is often below the text-matching tool's reporting threshold, thus causing the plagiarism to remain undetected. Our classification approaches for machine-paraphrased text can be a valuable complement to text-matching software. The additional analysis step could alert users when indicators of machine-obfuscated text have been identified.

We provide an illustrative example for text from arXiv, Wikipedia, and theses, their modified versions using SpinBot and SpinnerChief, and classification scores using text-matching software (Turnitin, PlagScan), the best performing neural language model (Longformer), and the best combination of machine learning classifier and word embeddings (w2v+NB) in \Cref{tab:showcase-examples}.

\begin{table*}[!t]
    \small
	\centering
    \caption{\label{tab:dataset_examples} An illustrative sample of three examples for each paraphrasing tool, data source, and classification method.} \label{tab:showcase-examples}
    \resizebox{\textwidth}{!}{
    	\begin{tabular}{p{10cm} c c c c c}
    		\toprule
            \textbf{Original Parapgraphs:}\\
            \multicolumn{6}{p{16cm}}{
            \begin{itemize}
                \vspace{-0.3cm}
                \item A mathematically rigorous approach to quantum field theory based on operator algebras is called an algebraic quantum field theory...
                \item "Nuts" contains 5 instrumental compositions written and produced by Streisand, with the exception of "The Bar", including additional writing from Richard Baskin. All of the songs were recorded throughout 1987...
                \item Most of activities are carried out internally using internal resources. The cost and financial demandingness of the project are presented in the table below...
            \end{itemize}} \\
            
    		\textbf{SpinBot Paraphrased} & \textbf{Source} & \textbf{Turnitin$^\dagger$} & \textbf{PlagScan$^\dagger$} & \textbf{ML$^*$} & \textbf{NLM$^*$} \\
    		\hline
            A \hl{numerically thorough way} to \hl{deal with} quantum field \hl{hypothesis dependent} on \hl{administrator} algebras is called an \hl{arithmetical} quantum field \hl{hypothesis}... & arXiv & 25.30 & 0.00 & 73.11 & 99.99\\
    		\\
    		\textbf{SpinnerChief-IF Paraphrased} & & & & \\
    		\hline
    		"Nuts" \hl{consists of five} instrumental compositions written and \hl{created} by Streisand, with the exception of "The Bar", \hl{which includes extra} writing from \hl{Rich} Baskin. All of the music were recorded \hl{in} 1987... & Wiki & 35.80 & 23.90 & 98.87 & 72.46\\
    		\\
    		\textbf{SpinnerChief-DF Paraphrased} & & & & \\
    		\hline
    		\hl{The majority} of activities are carried out internally \hl{making use of} internal resources. The cost and \hl{economic} demandingness of the project are \hl{illustrated} in the table below...  & Thesis & 0.00 & 0.00 & 49.81 & 63.81 \\
            
    		\bottomrule
    	\end{tabular}
    }
	\begin{tablenotes}[online]
	\item[\ding{227}] $^\dagger$text-match in \%.
	\item[\ding{227}] $^*$prediction score in \% for the best performing models Longformer and w2v+NB (see \Cref{sec.eval}).
	\item[\ding{227}] \hl{Red background} highlights changed tokens of the original version.
	\item[\ding{227}] Ellipsis (``...'') indicates the remainder of the paragraph.
	\end{tablenotes}
\end{table*}

%=====================================================================================
% Conclusion and Future Work
%=====================================================================================
\section{Conclusion}\label{sec.conclusion}
In this paper, we analyze two new collections (arXiv and theses), an additional paraphrasing tool (SpinnerChief), eight neural language models based on the Transformer architecture (\Cref{tab:class-results-nlm}), and two popular text-matching systems (Turnitin and PlagScan). We selected training and test sets that reflect documents particularly relevant for the plagiarism detection use case. The arXiv collection represents scientific papers written by expert researchers. Graduation theses of non-native English speakers provide writing samples of authors whose style varies considerably. Wikipedia articles represent collaboratively authored documents for many topics and one of the sources from which students plagiarize most frequently.  

We investigated the use of traditional, pre-trained word embedding models in combination with machine learning classifiers and recent neural language models. For eight of our nine test sets, Transformer-based techniques proposing changes in the training architecture achieved the highest scores. In particular, Longformer achieved the best classification performance overall.

Transferring the classifiers trained on the SpinBot training set to SpinnerChief test sets caused a drop in the approaches' average classification performance. For \spinnerchiefif, a test set exhibiting a similar ratio of altered words as the training set, the average F1-Micro scores of the best-performing classifiers dropped by approx. 21\%, that of human evaluators by approx. 13\%. However, the best-performing models were still capable of classifying machine-paraphrased paragraphs with F1-Micro scores ranging between 74.8\% to 80.5\%. We partially attribute the loss in performance in recognizing SpinnerChief test cases to the obfuscation's strength and not exclusively to deficiencies in the transferred classifiers.

We showed that our approaches can complement text-matching software, such as PlagScan and Turnitin, which often fail to identify machine-paraphrased plagiarism. The main advantage of machine learning models over text matching software is the models' ability to identify machine-paraphrased text even if the source document is not accessible to the detection system. The classification approaches we investigated could be integrated as an additional step within the detection process of text-matching software to alert users of likely machine-paraphrased text. The presence of such obfuscated text is a strong indicator of deliberate misconduct.

The classification approaches we devised are robust to identifying machine-paraphrased text, which educators face regularly. To support practitioners and facilitate an extension of the research on this important task, the data\footnote{\url{https://doi.org/10.5281/zenodo.3608000}} and code\footnote{\url{https://github.com/jpelhaW/ParaphraseDetection}} of our study, as well as a web-based demonstration system for the best-performing machine learning classifier\footnote{\url{http://purl.org/spindetector}} (NB+w2v) and neural language model\footnote{\url{https://huggingface.co/jpelhaw/longformer-base-plagiarism-detection}} (Longformer) are openly available.

\section{Future Work} \label{sec.future}

Our experiments indicate that obtaining additional training data is a promising approach for improving artificial intelligence-backed approaches for identifying machine-paraphrased text. Additional training data should cover more paraphrasing tools, topics, and languages. We see a community-driven open data effort as a promising option for generating a comprehensive training set. We encourage researchers investigating machine-paraphrase detection to share their data and contribute to the consolidation and extension of datasets, such as the one we publish with this paper.

Obtaining effective training data is challenging due to many paraphrasing tools, nontransparent paraphrasing approaches, frequent interconnections of paraphrasing tools, and the questionable business model of the tool providers. If paying paraphrasing services to obtain data proves prohibitive, a crowdsourcing effort could overcome the problem. Another interesting direction would be to use auto-encoding language models to paraphrase text or generate new text with auto-regressive models. This setup will be more realistic in the future as language models are publicly available and generate text that is difficult to distinguish from human writing.

\bibliographystyle{splncs04}
\bibliography{_bibliography.bib}

\end{document}